%% file: main.tex
\documentclass[10pt,twocolumn,letterpaper]{article}

\usepackage[pagenumbers]{iccv} 


\definecolor{iccvblue}{rgb}{0.21,0.49,0.74}
\usepackage[pagebackref,breaklinks,colorlinks,allcolors=iccvblue]{hyperref}
\usepackage{booktabs}       
\usepackage{array}
\usepackage{makecell}
\usepackage{multirow}
\usepackage{colortbl}
\usepackage{xcolor}   
\usepackage{caption} 
\usepackage{stfloats}
\usepackage{float}
\usepackage{fontawesome}
\usepackage{upgreek}
\definecolor{darkgrey}{rgb}{0.25, 0.25, 0.25} 
\usepackage{siunitx}
\usepackage{amssymb}
\def\paperID{6945} 
\def\confName{ICCV}
\def\confYear{2025}

\title{LoD-Loc v2: Aerial Visual Localization over Low Level-of-Detail City Models using Explicit Silhouette Alignment}

\author{Juelin Zhu$^{1}$, Shuaibang Peng$^{1}$, Long Wang$^{2}$, Hanlin Tan$^1$, Yu Liu$^1$, Maojun Zhang$^{1*}$, Shen Yan$^1$
\and
$^1$National University of Defense Technology, \ $^2$Westlake University\\
{\tt\small \{zhujuelin,psb24,hanlin\_tan,jasonyuliu,mjzhang,yanshen12\}@nudt.edu.cn, wanglongzju@gmail.com}
}

\begin{document}

\twocolumn[{%
    \renewcommand\twocolumn[1][]{#1}
    \maketitle 
    \begin{center}
        \centering
        \includegraphics[width=\textwidth]{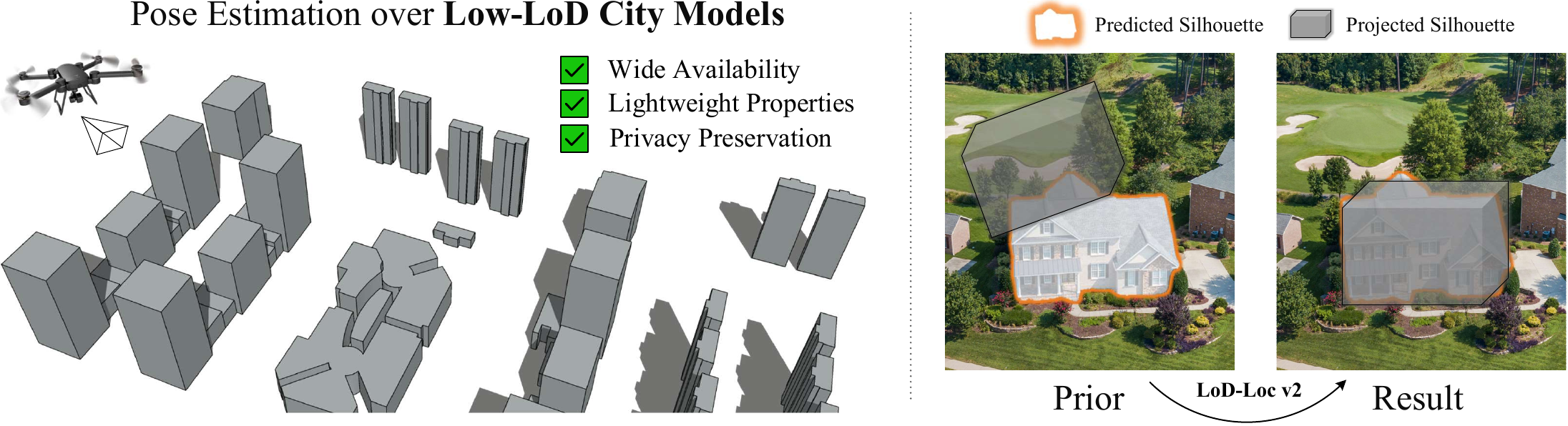}
        \captionof{figure}{In this paper, we introduce LoD-Loc v2 to tackle aerial visual localization using \textbf{low-LoD} city models. These models are characterized by wide availability, lightweight properties, and inherent privacy-preserving capabilities.
        Given a query image with its prior pose, our approach utilizes the explicit silhouette alignment to recover the camera pose. }
        \label{fig:intro}
    \end{center}%
}]
\renewcommand{\thefootnote}{*}
\footnotetext{Corresponding author}
\input{sec/0_abstract}    
\input{sec/1_intro}
\input{sec/2_related}
\input{sec/3_method}
\input{sec/4_Experiment.tex}
\input{sec/5_Conclusion.tex}
{
    \small
    \bibliographystyle{ieeenat_fullname}
    \bibliography{main}
}

\end{document}

%% file: sec/0_abstract.tex
\begin{abstract}

We propose a novel method for aerial visual localization over \textbf{low} Level-of-Detail (LoD) city models. Previous wireframe-alignment-based method LoD-Loc~\cite{zhu2024lod} has shown promising localization results leveraging LoD models. However, LoD-Loc mainly relies on high-LoD (LoD3 or LoD2) city models, but the majority of available models and those many countries plan to construct nationwide are low-LoD (LoD1). Consequently, enabling localization on low-LoD city models could unlock drones' potential for \textbf{global} urban localization. 
To address these issues, we introduce LoD-Loc v2, which employs a coarse-to-fine strategy using explicit silhouette alignment to achieve accurate localization over low-LoD city models in the air.
Specifically, given a query image, LoD-Loc v2 first applies a building segmentation network to shape building silhouettes. Then, in the coarse pose selection stage, we construct a pose cost volume by uniformly sampling pose hypotheses around a prior pose to represent the pose probability distribution.
Each cost of the volume measures the degree of alignment between the projected and predicted silhouettes. We select the pose with maximum value as the coarse pose. In the fine pose estimation stage, a particle filtering method incorporating a multi-beam tracking approach is used to efficiently explore the hypothesis space and obtain the final pose estimation. To further facilitate research in this field, we release two datasets with LoD1 city models covering 10.7 km$^2$, along with real RGB queries and ground-truth pose annotations. 
Experimental results show that LoD-Loc v2 improves estimation accuracy with high-LoD models and enables localization with low-LoD models for the first time. Moreover, it outperforms state-of-the-art baselines by large margins, even surpassing texture-model-based methods, and broadens the convergence basin to accommodate larger prior errors.
The project are available at \url{https://github.com/VictorZoo/LoD-Loc-v2}.

\end{abstract}

%% file: sec/1_intro.tex
\section{Introduction}
\label{sec:intro}



Aerial visual localization is a task that relies on visual inputs to estimate an Unmanned Aerial Vehicle (UAV) camera’s position and orientation relative to a georeferenced map.
This task plays a crucial role in the fields of computer vision and remote sensing, with applications in navigation~\cite{lu2018survey}, cargo transport~\cite{villa2020survey}, and surveillance~\cite{cai2022review, wu2024uavd4l}.

\input{table/dataset_table.tex}

Traditional aerial visual localization algorithms~\cite{chen2021real, wu2024uavd4l, yan2023render} typically rely on high-quality pre-constructed 3D georeferenced data~\cite{schonberger2016structure, panek2022meshloc, yan2023render, wu2024uavd4l}, which offer rich structure and color features for accurate representation. However, this introduces challenges such as difficulties in data acquisition, large storage requirements, and privacy concerns. 
Recently, LoD-Loc~\cite{zhu2024lod} proposed an aggressive pipeline that utilizes LoD city models as reference maps to perform aerial visual localization.
LoD models are 3D representations that focus solely on architectural structures~\cite{wysocki2024reviewing}, characterized by watertight geometry, object-level modeling, and hierarchical semantics across various LoDs, following the CityGML standard and its GML encoding~\cite{ groger2012ogc, kutzner2023ogc}.
Utilizing LoD city models for visual localization offers two advantages:
1) \textbf{Light-weight properties}: LoD models are highly compact, with their storage footprint in a given area being about $10^4$ times smaller than 3D texture models~\cite{zhu2024lod}, which alleviates storage burdens for edge deployments. 2) \textbf{Privacy preservation}: Since LoD city models only shape the basic 3D outlines of buildings in a highly abstract form, they lead to fewer concerns regarding privacy disclosure.

However, a limitation of the LoD-Loc~\cite{zhu2024lod} approach is its reliance on high-level LoD city models, such as LoD2 or LoD3, which provide enriched structural information necessary for its proposed neural wireframe alignment method.
Actually, in practical applications, the majority of available and open-source LoD models are of low detail (typically LoD1)~\cite{wysocki2024reviewing}, as summarized in \cref{Tab:table_dataset}. Notably, several countries, including the United States~\cite{open_city_model}, China~\cite{china_LoD1, SUN2024114057}, Switzerland~\cite{swisstopo, Swiss-LoD1}, Japan~\cite{Japan_LoD1}, Netherlands~\cite{3D_BAGS, Netherland-LoD1}, Singapore~\cite{SG-LoD1, SG-LoD2} and Germany~\cite{Das_3D}, have either invested in or are planning the development of nationwide LoD1 city models. Consequently, advancing research into visual localization methods that utilize low-LoD city models is pivotal for enhancing the \textbf{global} city localization capabilities of UAVs.



\begin{figure}[t]
  \centering
  \includegraphics[width=\linewidth]{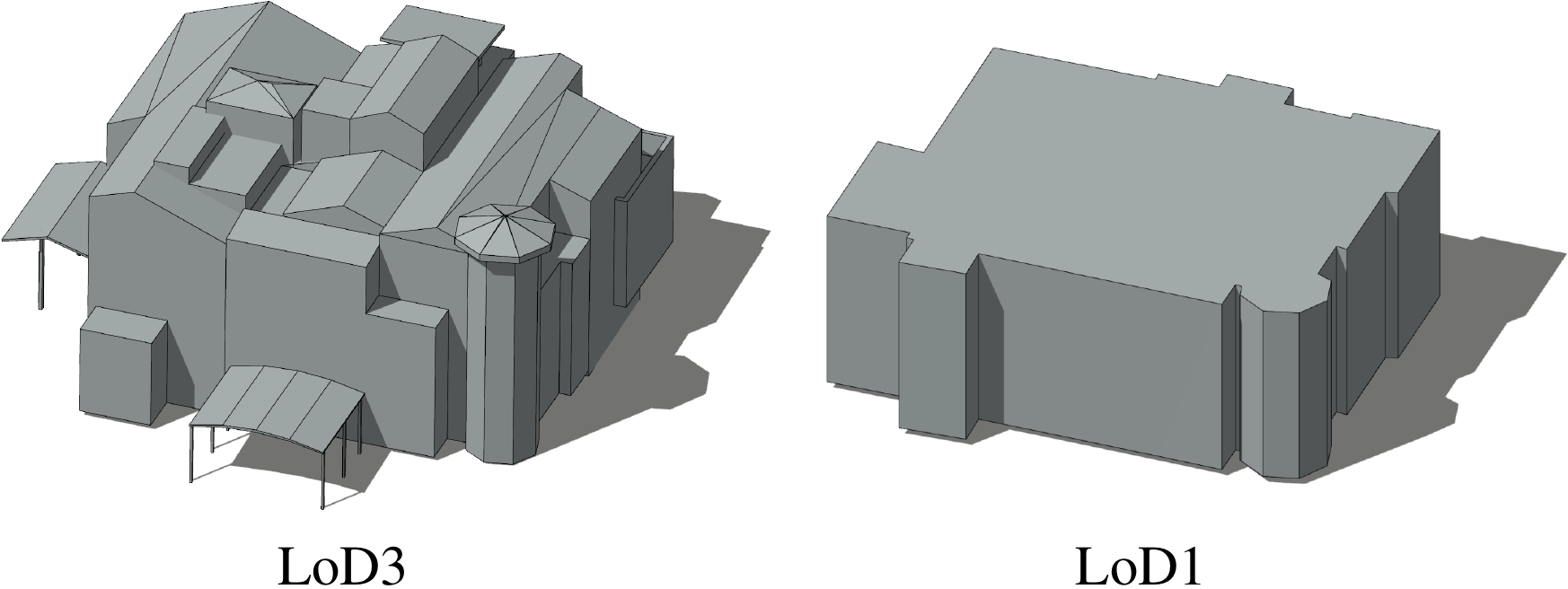}
  \caption{\textbf{Comparison between LoD3 and LoD1.} The LoD3 model provides a highly accurate representation of the building's geometry. The LoD1 model focuses on the building's basic shape without intricate details. Consequently, achieving effective visual localization using low-LoD city models poses a significant challenge. More details can be found in the Appendix.}
  \label{fig:1_3vs3_0}
\end{figure}

This task is quite challenging, as shown in \cref{fig:1_3vs3_0}, LoD1 models not only lack textured features but also lose significant structural details compared to high-LoD models. In this work, we explore a straightforward yet effective idea: even when wireframe representations derived from LoD1 models can not achieve precise alignment, the silhouettes remain amenable to accurate matching when the pose is correctly estimated as shown in \cref{fig:intro}.
Building on this insight, we introduce LoD-Loc v2, a novel approach tailored for visual localization using low-LoD city models.
Specifically, we first extract the predicted silhouettes of the query image using a segmentation network and then propose a two-stage method to gradually refine the alignment between the predicted and projected silhouettes. Given the real built-in sensor prior pose data, the coarse pose selection stage fixes the 2-DoF orientation (yaw and pitch) and generates pose hypotheses uniformly in the 3 degrees of freedom (DoF) translation and the yaw direction. Based on the generated pose hypotheses, real-time rendering technique is applied to project the LoD model onto a 2D plane, yielding the projected building silhouettes. We then score each pose hypothesis by the alignment between the projected and predicted silhouettes, thereby forming a 4D pose cost volume. The hypothesis with the highest score from the volume is selected as the coarse pose. To further enhance pose accuracy, the fine pose estimation stage employs a particle filter-based optimizer~\cite{kwon2007particle, thrun2002probabilistic} that iteratively explores a range of candidate hypotheses within the solution space, leading to the accurate pose recovery.




To facilitate research in this field, we release two new datasets equipped with LoD1 city models from distinct regions. Both datasets comprise drone-captured query images with priors and accurate pose annotations, spanning a total area of 10.7 km$^2$.
We conduct comprehensive experiments on both the LoD1 dataset and the combined LoD2/LoD3 datasets~\cite{zhu2024lod}. The results demonstrate that our proposed method enables UAV localization using low-LoD city models, outperforming the LoD-Loc~\cite{zhu2024lod} method by a large margin and even surpassing the current state-of-the-art (SOTA) approach based on texture models~\cite{wu2024uavd4l, panek2023visual, trivigno2024unreasonable}. Simultaneously, our method expands the convergence basin to handle larger prior errors, making it a potential solution for scenarios with GPS-constrained situations.




In summary, our main contributions include:
\begin{enumerate}
    \item We introduce a novel explicit-silhouette-alignment based framework, LoD-Loc v2, enabling aerial visual localization with low-LoD city models for the first time.
    \item We release two geo-tagged LoD1 datasets covering 10.7 km$^2$, designed to facilitate future research.
    \item Extensive experiments demonstrate that LoD-Loc v2 achieves outstanding localization performance in both low and high-LoD models, while also expanding the convergence basin to accommodate larger prior errors.
\end{enumerate}

%% file: table/dataset_table.tex
\begin{table}[t]
\centering
\setlength{\tabcolsep}{8pt} 
\begin{tabular}{lccc}
\toprule
Name           & LoD1 & LoD2 & LoD3 \\ \hline
Project PLATEAU~\cite{seto2023role} & {\cellcolor[rgb]{0.941,0.941,0.941}}\checkmark{}       & {\cellcolor[rgb]{0.902,0.902,0.902}}\checkmark{}       & {\cellcolor[rgb]{0.851,0.851,0.851}}\checkmark{}       \\
City JSON~\cite{City_JSON}       & {\cellcolor[rgb]{0.941,0.941,0.941}}\checkmark{}       & {\cellcolor[rgb]{0.902,0.902,0.902}}\checkmark{}       &         \\
Netherlands~\cite{peters2022automated}     & {\cellcolor[rgb]{0.941,0.941,0.941}}\checkmark{}       & {\cellcolor[rgb]{0.902,0.902,0.902}}\checkmark{}       &        \\
Swiss-TOPO~\cite{swisstopo}      & {\cellcolor[rgb]{0.941,0.941,0.941}}\checkmark{}     & {\cellcolor[rgb]{0.902,0.902,0.902}}\checkmark{}      &         \\
Open City Model~\cite{open_city_model}       & {\cellcolor[rgb]{0.941,0.941,0.941}}\checkmark{}       &         &         \\
Das 3D-Modell~\cite{Das_3D}   & {\cellcolor[rgb]{0.941,0.941,0.941}}\checkmark{}      &        &       \\
East Asia~\cite{shi2024last}      & {\cellcolor[rgb]{0.941,0.941,0.941}}\checkmark{}       &        &    \\
GABLE~\cite{SUN2024114057} & {\cellcolor[rgb]{0.941,0.941,0.941}}\checkmark{}       &         &        \\ \bottomrule
\end{tabular}
\caption{\textbf{Available LoD models in different datasets.} LoD1 data is widely available across most regions, whereas LoD2 and LoD3 data are restricted to specific areas, underscoring the necessity to explore localization techniques using low-LoD models.}
\label{Tab:table_dataset}
\end{table}

%% file: sec/2_related.tex
\section{Related Works}
\label{sec: Ralted works}


\noindent \textbf{Localization over information-rich 3D map.} 
Conventionally, most existing methods~\cite{wu2024uavd4l, sun2021loftr, sarlin2020superglue, sarlin2019coarse, panek2022meshloc} tackle pose estimation by establishing correspondences between the query image and sparse SfM~\cite{schonberger2016structure} or mesh~\cite{panek2022meshloc} reconstruction of the scene. The visual localization process utilizing the above 3D references can be summarized as follows: given a query image, image retrieval techniques \cite{arandjelovic2016netvlad,ge2020self} are first employed to identify reference images that share similar viewpoints. Next, feature matching algorithms \cite{lowe2004distinctive,detone2018superpoint,karkus2018particle,sarlin2020superglue,sun2021loftr} establish accurate 2D correspondences between the query image and the identified reference images, converting these correspondences into 3D relationships by incorporating depth information. Finally, pose estimation is conducted using the PnP-RANSAC algorithm \cite{haralick1994review,kneip2011novel,barath2019magsac,chum2008optimal,fischler1981random,chum2003locally,yan2023long}. Despite the impressive capabilities of these 3D models in enhancing localization accuracy, they are difficult to reconstruct and maintain. Moreover, their substantial size requires considerable storage and computational resources, while their detailed information raises privacy and data protection concerns.

\noindent \textbf{Localization over 2D map.} 
In addressing the issue of map reconstruction and maintenance, some works introduce taking satellite images as references~\cite{shi2022beyond, shi2020looking, xia2022visual, zheng2020university, dai2021transformer, SDPL_shift}. The workflow of these methods generally involves constructing a georeferenced image database, dividing the target area into uniform tiles, retrieval by global descriptors~\cite{galvez2012bags,arandjelovic2013all,arandjelovic2016netvlad}, re-ranking to select the best matches~\cite{barbarani2023local,hausler2021patch,zhu2023r2former,lu2024towards}, and projection calculation to estimate the precise location of the query image~\cite{moskalenko2024visual}. In the context of privacy, some methods are introduced to utilizing OpenStreetMap~\cite{sarlin2023orienternet, wu2024maplocnet} or implicit neural map\cite{sarlin2024snap} as a reference. However, due to the lack of altitude information, these methods can estimate at most a 3-DoF pose, including planar position and heading.



\noindent \textbf{Localization over LoD city model.}
Recently, LoD-Loc \cite{zhu2024lod} proposes to leverage LoD models as the cue, which aligns LoD models projected line segments with wireframes predicted by neural networks to achieve precise pose estimation. The primary advantage of the LoD-Loc method lies in its independence from complex 3D representations, enabling efficient and accurate visual localization using LoD models. 
However, the LoD-Loc method advocates localization on high-detail models at LoD3 or LoD2, which are limited in availability, thus restricting its scalability for global applications. Besides, since most of open-source LoD models are at low-LoD, it is essential to develop aerial visual localization algorithms based on low-detail models.
Therefore, this paper proposes a novel visual localization method, aiming to enhance the generality and practicality of localization systems by using widely available low-LoD models.

\noindent \textbf{Image segmentation.} 
Segment Anything Model (SAM) \cite{kirillov2023segment} introduces a prompt-based image segmentation task aimed at outputting effective segmentation masks given input prompts, such as bounding boxes or points indicating areas of interest. As segmentation models have evolved, HQ-SAM \cite{ke2024segment} has improved the segmentation quality of the SAM model by incorporating high-quality output tokens and a global-local feature fusion strategy. Additionally, FastSAM \cite{zhao2023fast}, MobileSAM \cite{zhang2023faster}, and EfficientSAM \cite{xiong2024efficientsam} have enhanced the convenience and efficiency of the SAM model for real-time applications by optimizing the model architecture and computational efficiency. Furthermore, SAM2~\cite{ravi2024sam} has strengthened the adaptability of model to dynamic environments by integrating a streaming memory architecture and multi-sensor data fusion techniques, thereby facilitating broader applications in various fields, including medical imaging \cite{ma2024segment,deng2023segment,mazurowski2023segment,wu2023medical}, motion segmentation \cite{xie2024moving}, camouflage target detection \cite{tang2023can}, and building extraction \cite{chen2024rsprompter,ren2024segment}. Inspired by these advancements, our approach first extracts building silhouettes using a segmentation network and then refines the pose to align the extracted silhouettes with the projected building silhouettes.

%% file: sec/3_method.tex
\section{Method}
\label{sec: Method}


\noindent \textbf{Problem formulation.}
In typical localization tasks, the goal is to estimate the absolute 6-DoF pose of an image in the real world. In the UAV system, due to the gravity direction is provided by the onboard inertial sensors~\cite{yan2023long, sarlin2023orienternet, sarlin2024snap, lynen2020large, Fragoso_2020_CVPR}, the problem can be simplified to estimating a 4-DoF pose ${\boldsymbol{\xi}}^* = (x, y, z, \theta)$, where $(x, y, z) \in \mathbb{R}$ denotes the translation and $\theta \in [0, 2\pi)$ is the yaw angle. 

\begin{figure*}[ht]
  \centering
  \includegraphics[width=\linewidth]{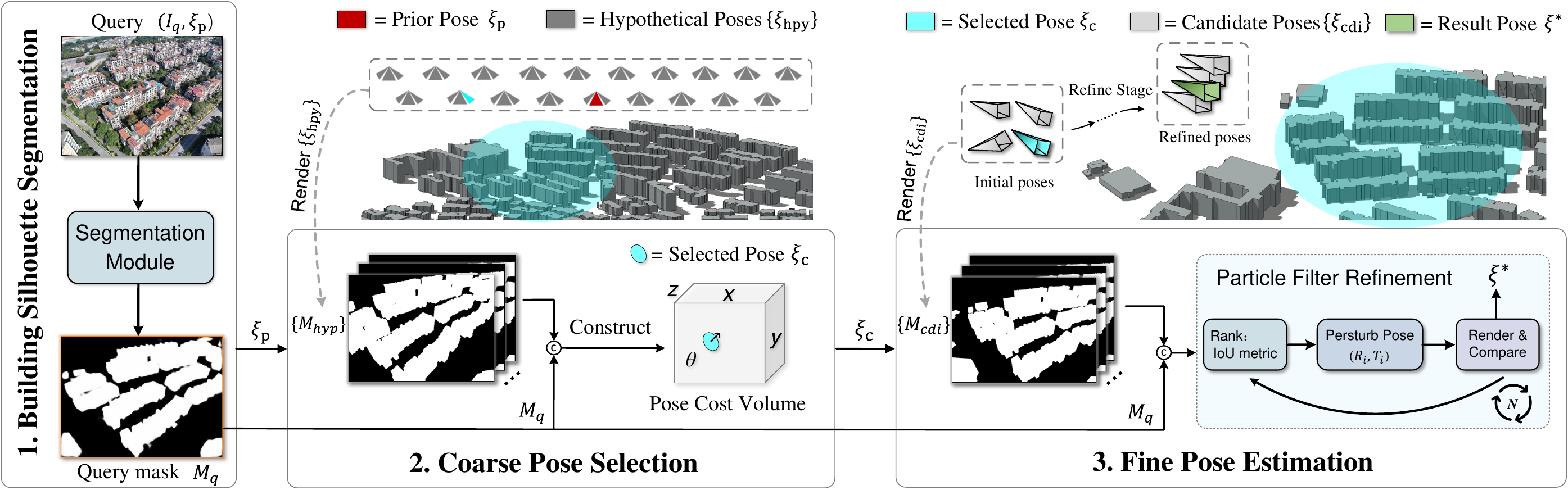}
  \caption{\textbf{Overview of LoD-Loc v2.} 1. LoD-Loc v2 employs a building segmentation module to extract building silhouettes $M_q$ from the query image $I_q$ (\cref{sec:building_seg}). 2. A 4D pose cost volume $\mathcal{C}$ is built for pose hypotheses $\{ \boldsymbol{\xi}_{hyp} \}$ sampled around the prior pose ${\boldsymbol{\xi}}_p$ to select the pose ${\boldsymbol{\xi}}_c$ with the highest probability, based on the alignment between projected and predicted building silhouettes~(\cref{sec:pose_selection}). 3. A particle filter refinement is applied to refine the pose ${\boldsymbol{\xi}}_c$ to obtain a final accurate pose ${\boldsymbol{\xi}}^{*}$ (\cref{sec:filter_refine}).}
  \label{fig:pipeline}
\end{figure*}

\noindent \textbf{Overview - \cref{fig:pipeline}.}
Given a LoD model $\mathcal{M}$, a query image $I_q$ and its prior pose estimate ${\boldsymbol{\xi}}_p$, the proposed LoD-Loc v2 aims to determine the precise absolute pose ${\boldsymbol{\xi}}^*$. 

\subsection{Building Silhouette Segmentation}
\label{sec:building_seg}

We develop a SAM2-based U-like network~\cite{xiong2024sam2} to extract building silhouettes $M_q$ from the $I_q$. The network consists of SAM2 encoders, U-Net decoders, receptive field blocks (RFBs), and adapters for efficient fine-tuning~\cite{liu2018receptive}. Specifically, given an input query image \( I_q \in \mathbb{R}^{3 \times H \times W} \), the encoder block outputs four levels of features \( F_i^e \in \mathbb{R}^{C_i \times \frac{H}{2^{i+1}} \times \frac{W}{2^{i+1}}} \) for \( i \in \{1,2,3,4\} \), where \( C_i \in \{144, 288, 576, 1152\} \). For each \( F_i^e \), the RFB reduces the channel count to 64,  thereby yielding in \( F_i^{rbf} \in \mathbb{R}^{64 \times \frac{H}{2^{i+1}} \times \frac{W}{2^{i+1}}} \) for \( i \in \{1,2,3,4\} \). Subsequently, the decoder blocks in the U-like structure perform sequential upsampling, incorporating each feature layer $F_i^{rbf}$. The output features of each decoder block then pass through a \( 1 \times 1 \) convolutional segmentation head, producing a corresponding segmentation result \( S_i \) for \( i \in \{1,2,3\} \). Finally, we apply a sigmoid activation and normalization to $S_1$, then threshold it at $\delta_s$ to produce the final query mask $M_q$. The detailed model architecture can be found in the Appendix.

\subsection{Coarse Pose Selection}
\label{sec:pose_selection}

We sample pose hypotheses $\{ \boldsymbol{\xi}_{hyp} \}$ around the sensor prior pose ${\boldsymbol{\xi}}_p$ to construct a pose cost volume $\mathcal{C}$ by evaluating the alignment between the projected silhouettes $\{ M_{hyp} \}$ and predicted silhouettes $M_q$. The pose ${\boldsymbol{\xi}}_c$ with the highest probability in the cost volume is then selected for the next fine pose estimation stage. 

Specifically, we first decouple the prior sensor pose ${\boldsymbol{\xi}}_p$ into six degrees, known as $(x_p, y_p, z_p, \theta_p, \rho_p, r_p)$, where $(x_p, y_p, z_p)$ denote the translation in 3D space and $(\theta_p, \rho_p, r_p)$ represents the Euler angles, namely yaw, pitch, roll. Since the errors in the gravity direction of device sensors are negligible, we then fix the rotation axis $(\rho_c, r_c = \rho_p, r_p)$ and solely focus on the remaining 4-DoF (i.e., (\( x, y, z, \theta_p\))) to construct the pose cost volume $\mathcal{C}$.
Next, we uniformly sample within the 4-DoF space centered at the prior pose ${\boldsymbol{\xi}}_p$, with a sampling range of $\left \{ \mathrm{r}(d)\right \}$ and a sampling count of $\left \{ \mathrm{n}(d) \right \}$. The sampled poses $\{ {\boldsymbol{\xi}}_{hyp} \}$ are generated as:
\begin{equation}
    \label{eq:pose_hyp}
    \{ \boldsymbol{\mathcal{\xi}}_{hyp}(d) \} = \left\{ d_p - \mathrm{r}(d)/{2} + k \cdot \Delta s \right\}
\end{equation}
where $d \in (\mathrm{x}, \mathrm{y}, \mathrm{z}, \uptheta)$ means one of the axes, $k = \{ 0, 1, \dots, \mathrm{n}(d) - 1 \}$ is the index for each offset from the starting position, and $\Delta s = \frac{\mathrm{r}(d)}{\mathrm{n}(d) - 1}$ represents the stride. 

Based on each hypothetical pose $\boldsymbol{\xi}_{hyp}$, we utilize real-time rendering technology OpenSceneGraph ($\mathrm{OSG}$)~\cite{OSG} to project the LoD model onto a 2D plane, generating a synthetic silhouette image \(  I_{hyp}  \). Subsequently, a threshold of $\delta_I$ is applied to \( \{ I_{hyp} \} \) to produce binary masks $\{ M_{hyp} \}$.We define a silhouette alignment cost for a query image $I_q$ across each hypothetical pose as:
\begin{equation}
    \label{eq:cost}
    \mathrm{c}(I_q, I_{hyp}) = \frac{|M_q \cap M_{hyp}|}{|M_q \cup M_{hyp}|}
\end{equation}
where \( M_q \cap M_{hyp} \) means the intersection region of the predicted and projected mask, \( M_q \cup M_{hyp} \) is the union region of the predicted and projected mask, and \( | \cdot | \) indicates the number of pixels in the set.

By combining all hypothetical costs in a grid manner across the four dimensions ($\mathrm{x} \), \( \mathrm{y} \), \( \mathrm{z}, \uptheta$), we obtain a pose cost volume $ \mathcal{C}(I_q, I_{hyp}) =  \{ \mathrm{c}(I_q, I_{hyp}^i) \mid i \in \mathrm{n}(d)  \} $. Finally, we use an $argmax$ operation upon $\mathcal{C}$ to select the coarse pose ${\boldsymbol{\xi}}_c$ with maximum value.

\input{table/table1.tex}

\subsection{Fine Pose Estimation}
\label{sec:filter_refine}

To further refine the coarse pose ${\boldsymbol{\xi}}_c$, inspired by~\cite{karkus2018particle, lin2023parallel, maggio2023loc, humenberger2022investigating, kendall2015posenet, trivigno2024unreasonable, Pietrantoni_2023_CVPR}, we adopt a particle filtering approach in the refinement stage to estimate the system state based on observations and dynamic adjustments. 

\noindent \textbf{Preliminary.} 
The basic concept of particle filters is that, given an initial hypothesis, we can obtain new state hypotheses by perturbing the initial state, then evaluate the cost function and iteratively optimize the estimate. In a typical pose estimation task, the camera pose $\boldsymbol{\xi}$ is selected as the state variable. Starting from the initial state \( {\boldsymbol{\xi}}_c = (\mathbf{R}_c, \mathbf{t}_c) \), where \( \mathbf{t}_c \in \mathbb{R}^3 \) represents the camera center and \( \mathbf{R}_c \in \mathbb{R}^4 \) is a quaternion that provides a stable rotation framework~\cite{lepetit2005monocular}, it applies perturbations \( \delta \) to generate a set of new candidate poses \( \{ {\boldsymbol{\xi}}_{cdi} \} \). These candidate poses lie on the manifold of \( \text{SO}(3) \times \text{T}(3) \), representing the combined rotation and translation spaces. They are subsequently evaluated and incrementally refined using the predefined loss function $\mathcal{L}(\cdot)$, 
thereby transforming the problem into the following optimization:
\begin{equation}
    \label{eq:opt}
    \boldsymbol{\xi}^* = \operatorname*{argmin}_{ \{ \boldsymbol{\xi}_{cdi} \} \in \mathrm{SO}(3) \times \mathrm{T}(3)} \mathcal{L}(\boldsymbol{\xi}_{cdi} \mid I_q, I_{\boldsymbol{\xi}_{cdi}})
\end{equation}
where $ I_{\boldsymbol{\xi}_{cdi}}$ is the rendered candidate.

\noindent \textbf{Our approach.} 
In our case, the particle filter models the posterior distribution of the coarse pose \( {\boldsymbol{\xi}}_c \) as \( p({\boldsymbol{\xi}}_c\mid{\boldsymbol{Z}}_i) \), where the particle set \( {\boldsymbol{Z}}_i = \{(\xi_i^1, \pi_i^1), \dots, (\xi_i^n, \pi_i^n)\} \) consists of multiple particles and their corresponding weights \( \pi \). Each particle’s weight \( \pi_i^j \) represents its sampling probability, estimated based on the \cref{eq:cost} and \cref{eq:opt}, i.e., $\mathcal{L}(\boldsymbol{\xi} \mid I_q, I_{\boldsymbol{\xi}})=\mathcal{C}(I_q, I_{\boldsymbol{\xi}})$.  
Since these particle states \( \{ \xi_i^1, \dots, \xi_i^n \} \) are defined on the \( \text{SO}(3) \times \text{T}(3) \) manifold, they can be perturbed using Lie algebra~\cite{chiuso2000monte, choi2012robust, kwon2010visual}. Since errors along the pitch and roll axes are negligible, we keep these parameters fixed $(\rho^*, r^* = \rho_c, r_c)$ and apply rotational perturbations only around the yaw axis within the \( \text{SO}(3) \) subspace, while introducing translational perturbations along the \( ( \mathrm{x} \), \( \mathrm{y} \), \( \mathrm{z} ) \) axes in \( \text{T}(3)  \) space. Additionally, to explore the hypothesis space efficiently, we adopt a multi-beam hypothesis tracking approach~\cite{choi2012robust, trivigno2024unreasonable}. Each beam represents an independent optimization thread, allowing parallel optimization tasks to increase the likelihood of movement in the correct direction.

\subsection{Supervision}
\label{sec:supervision}

The LoD-Loc v2 is trained in a supervised manner from pairs of query images $I_q$ and ground-truth building masks \( G \).
We employ the loss function $L_{\text{seg}}$ to supervise the building silhouette segmentation network. 
The $L_{\text{seg}}$ consists of a Binary Cross-Entropy (BCE$_{\text{w}}$) loss and a weighted Intersection over Union (IoU$_{\text{w}}$) loss~\cite{fan2020pranet, wei2020f3net}, described as: 
\begin{equation}
    \label{eq:seg_loss}
    L_{\text{seg}} = \sum_{i=0}^{2} \left ( \text{BCE}_{\text{w}}[S_i, G] + \text{IoU}_{\text{w}}[S_i,G] \right )_{mean}
\end{equation}

Notice that, during the training phase, the SAM2 encoder is frozen to reduce the number of trainable parameters, thus lowering computational and storage costs. 

%% file: table/table1.tex
\begin{table*}[ht]
\centering
\setlength{\tabcolsep}{8.5pt} 
\begin{tabular}{lccccccccc} 
\toprule
\multicolumn{2}{c}{\multirow{2}{*}{Method}}                 & \multicolumn{4}{c}{\textit{in-Traj.}}                     & \multicolumn{4}{c}{\textit{out-of-Traj.}}                 \\ \cmidrule(r){3-6} \cmidrule(r){7-10} 
\multicolumn{2}{c}{}                                        & 2m-2°           & 3m-3°           & 5m-5°   &T.e./R.e.        & 2m-2°           & 3m-3°           & 5m-5°   &T.e./R.e.        \\ \cmidrule(r){1-10}
\multicolumn{2}{c}{Prior}                            & 0              & 0              &  4.3    & 6.48/1.63        & 0              & 0              & 0.36     & 11.1/0.92      \\ \hline
\rowcolor[rgb]{1,0.992,0.980} {\cellcolor[rgb]{1,0.992,0.980}}                          & SIFT+NN        & 73.13 & 78.62 & 80.42  & 1.13/044     & 82.39 & 85.13  & 86.36  & 0.87/0.29                                        \\
\rowcolor[rgb]{1,0.992,0.980} {\cellcolor[rgb]{1,0.992,0.980}}                          & SPP+SPG                     & 91.71 & 92.02 & 92.14   & 0.79/0.29    & 93.43 & 93.70  & 93.80   & 0.74/0.19                                       \\
\rowcolor[rgb]{1,0.992,0.980} {\cellcolor[rgb]{1,0.992,0.980}}                          & LoFTR                     & 84.98 & 88.09 & 88.90 & 0.81/0.29    & 91.56 & 92.02  & 92.11   & 0.79/0.20                                    \\
\rowcolor[rgb]{1,0.992,0.980} {\cellcolor[rgb]{1,0.992,0.980}}                          & e-LoFTR                    & 84.47 & 88.21 & 88.96  & 0.96/0.35   & 91.06 & 91.93  & 92.02   &0.86/0.22                                     \\
\rowcolor[rgb]{1,0.992,0.980} \multirow{-5}{*}{{\cellcolor[rgb]{1,0.992,0.980}}\begin{tabular}[c]{@{}>{\cellcolor[rgb]{1,0.992,0.980}}c@{}}UAVD4L~\cite{wu2024uavd4l}\\ \textit{Texture model}\end{tabular}} & RoMA                      & 93.27 & 93.70 & 93.77 & 0.78/0.25    & 95.03 & 95.53  & 95.53   &0.73/0.22                                     \\  \hline
\rowcolor[rgb]{0.941,0.941,0.941} {\cellcolor[rgb]{0.941,0.941,0.941}}                          & SIFT+NN                     & 0                                         & 0                                         & 0        &-                             & 0                                         & 0                                         & 0           &-                           \\
\rowcolor[rgb]{0.941,0.941,0.941} {\cellcolor[rgb]{0.941,0.941,0.941}}                          & SPP+SPG                     & 0                                         & 0                                         & 0         &-                             & 0                                         & 0                                         & 0           &-                          \\
\rowcolor[rgb]{0.941,0.941,0.941} {\cellcolor[rgb]{0.941,0.941,0.941}}                          & LoFTR                     & 0                                         & 0                                         & 0          &-                            & 0                                         & 0                                         & 0             &-                        \\
\rowcolor[rgb]{0.941,0.941,0.941} {\cellcolor[rgb]{0.941,0.941,0.941}}                          & e-LoFTR                    & 0                                         & 0                                         & 0           &-                         & 0                                         & 0                                         & 0            &-                         \\
\rowcolor[rgb]{0.941,0.941,0.941} \multirow{-5}{*}{{\cellcolor[rgb]{0.941,0.941,0.941}}\begin{tabular}[c]{@{}>{\cellcolor[rgb]{0.941,0.941,0.941}}c@{}}CAD-Loc~\cite{panek2023visual}\\\textit{LoD model}\end{tabular}} & RoMA                      & 0                                         & 0                                         & 0            &-                             & 0                                         & 0                                         & 0          &-                             \\   \hline
\rowcolor[rgb]{0.902,0.902,0.902} {\cellcolor[rgb]{0.902,0.902,0.902}}                                                                                                                    & DINOv2                      & 1.20                                         & 4.10                                         & 17.40                    &  8.29/2.58                 & 2.40                                         & 7.40                                        & 26.10              &  7.02/2.29                            \\
\rowcolor[rgb]{0.902,0.902,0.902} \multirow{-2}{*}{{\cellcolor[rgb]{0.902,0.902,0.902}}\begin{tabular}[c]{@{}>{\cellcolor[rgb]{0.902,0.902,0.902}}c@{}}MC-Loc~\cite{trivigno2024unreasonable}\\\textit{LoD model}\end{tabular}} & RoMa                     & 0.10                                         & 0.60                                         & 3.30            &  10.6/8.60                         & 0.20                                          & 0.90                                         & 3.30     &  16.9/3.88                                  \\  \hline
\rowcolor[rgb]{0.851,0.851,0.851} \begin{tabular}[c]{@{}>{\cellcolor[rgb]{0.851,0.851,0.851}}c@{}}LoD-Loc~\cite{zhu2024lod}\\\textit{LoD model}\end{tabular} & -                                                           & 49.56                                 & 71.82                                 & 89.09              &  3.32/1.48             & 54.20                                 & 75.05                        & 89.51    &  3.33/1.18                          \\ 
\hline
\rowcolor[rgb]{0.898,0.898,1} {\cellcolor[rgb]{0.898,0.898,1}}                                                                                                                          & no refine              & 2.99           & 7.73           & 27.87    & 6.19/0.67   & 11.68          & 29.88           & 51.14    & 4.78/0.92     \\
\rowcolor[rgb]{0.898,0.898,1} {\cellcolor[rgb]{0.898,0.898,1}}                                                                                                                          & no select              & 93.50          & 98.40          & 99.50  & 0.74/0.17     & 90.50          & 94.80           & 96.90     & 0.77/0.16       \\
\rowcolor[rgb]{0.898,0.898,1} \multirow{-3}{*}{{\cellcolor[rgb]{0.898,0.898,1}}\begin{tabular}[c]{@{}>{\cellcolor[rgb]{0.898,0.898,1}}c@{}}\textbf{LoD-Loc v2}\\\textit{LoD model}\end{tabular}} & \textbf{Full}          & \textbf{93.70} & \textbf{98.40} & \textbf{99.50} & \textbf{0.72/0.15} & \textbf{97.90} & \textbf{99.80} & \textbf{100.00}  & \textbf{0.71/0.14} \\
\hline
\end{tabular}
\caption{\textbf{Quantitative comparison results of different methods over UAVD4L-LoDv2 dataset.} T.e. and R.e. denote median translation error (m) and median rotation error (°), respectively,}
\label{Tab:table_1}
\end{table*}

%% file: sec/4_Experiment.tex
\section{Experiments}
\label{sec: Experiments}

\input{table/table_swiss.tex}

Extensive experiments are conducted on the LoD1 datasets, as well as on the LoD3 and LoD2 datasets~\cite{zhu2024lod}, to validate the effectiveness of our proposed model.

\noindent \textbf{Datasets.}
We release two new datasets, namely UAVD4L-LoDv2 and Swiss-EPFLv2. The UAVD4L-LoDv2 dataset covers a 2.5 km$^2$ area with georeferenced LoD1 models, which was automatically generated from the textured mesh model of the UAVD4L~\cite{wu2024uavd4l} by the DP modeler~\cite{DPmodeler}. The Swiss-EPFLv2 provides a semi-automatically constructed LoD1 model, covering an 8.2 km$^2$ area. The queries in both datasets are derived from the UAVD4L-LoD and Swiss-EPFL datasets, respectively~\cite{zhu2024lod}.
Besides, the experiments also utilized the UAVD4L-LoD and Swiss-EPFL datasets, which contain semi-automatically generated LoD3 and LoD2 models, respectively, along with drone-captured query images and pose annotations. Detailed information is provided in the Appendix and \cite{zhu2024lod}.

\noindent \textbf{Baseline.}
We compare our proposed method with visual localization baselines from two categories: UAVD4L~\cite{wu2024uavd4l}, which performs localization on texture models; CAD-Loc~\cite{panek2023visual}, MC-Loc~\cite{trivigno2024unreasonable}, and LoD-Loc~\cite{zhu2024lod}, which operate on LoD models. Both UAVD4L and CAD-Loc employ keypoint-based strategies: 1) the SIFT~\cite{lowe2004distinctive} descriptor with traditional Nearest Neighbor (NN) matching, 2) learning-based extractor SuperPoint (SPP)~\cite{detone2018superpoint} with Superglue (SPG)~\cite{sarlin2020superglue}, 3) detector-free matcher LoFTR~\cite{sun2021loftr} with its variant 4) e-LoFTR~\cite{wang2024efficient}, and 5) dense feature matcher RoMa~\cite{edstedt2024roma}. Additionally, MC-Loc utilizes two backbones as its feature extraction encoder: 6) a pre-trained DINOv2~\cite{oquab2023dinov2} encoder, and 7) the RoMa~\cite{edstedt2024roma} dense feature extractor. 8) LoD-Loc~\cite{zhu2024lod} is implemented with its default parameters. Further details on the implementation of the baseline experiments can be found in the Appendix.

\noindent \textbf{Metrics.}
We follow standard localization evaluation metrics~\cite{toft2020long} and set thresholds at $(2m, 2^{\circ})$, $(3m, 3^{\circ})$, and $(5m, 5^{\circ})$, as same as the LoD-Loc~\cite{zhu2024lod}.

\subsection{Implementation Details}
During the training phase, the dataset for the segmentation module is derived from a subset of the synthetic images from the UAVD4L database~\cite{wu2024uavd4l}, excluding images without buildings. To further enrich the training data, additional synthetic images are generated based on~\cite{wu2024uavd4l} rendering method, with pitch angles set to $30^{\circ}$ or $60^{\circ}$.
The ground truth mask labels $G$, are generated by projecting the LoD model~\cite{zhu2024lod} onto a 2D plane with paired poses.
In total, $18,395$ images of training data were generated. The segmentation module accepts input images of size \( (1024, 1024) \). To optimize training, the learning rate is set to $0.001$, and the AdamW optimizer is adopted with a weight decay of \( 5 \times 10^{-4} \) to prevent overfitting. The number of training epochs is set to $20$ to ensure sufficient learning. In this process, the SAM2 encoder's pre-trained model is configured as \texttt{hiera\_large}.


\begin{figure*}[th]
    \centering
	\includegraphics[ width=1.0\linewidth]{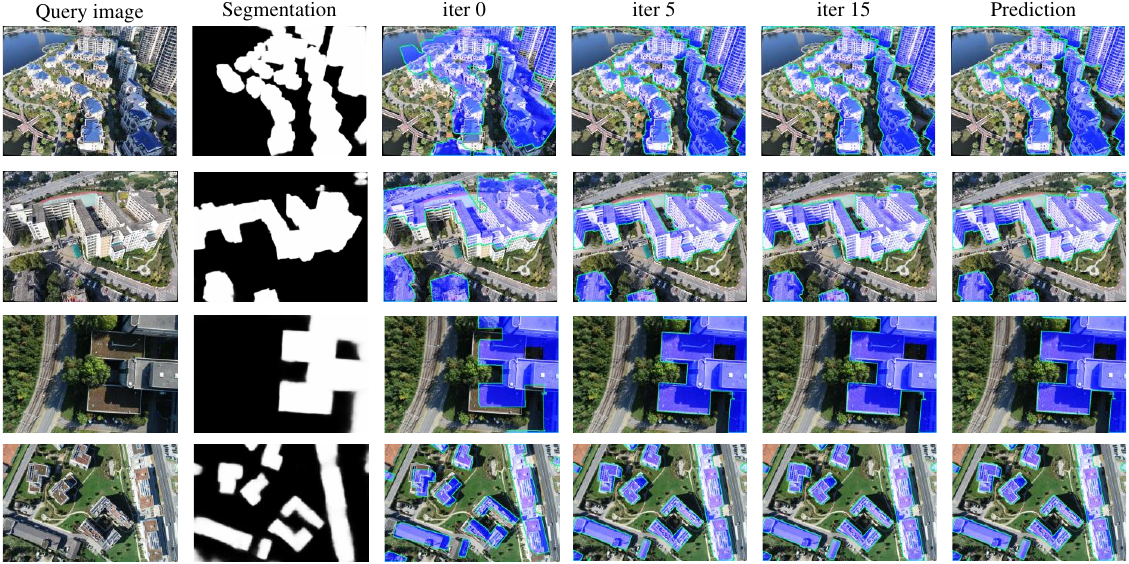}
	\caption{\textbf{Visualization of segmentation results and overlay of different iterations.} The visualized segmentation results demonstrate the model’s excellent segmentation performance. Better alignment indicates a more accurate pose prediction. Besides, the top two rows are from the UAVD4L-LoDv2 dataset, while the bottom two rows are from the Swiss-EPFLv2 dataset.
}
    \label{fig:exp_seg}
\end{figure*}

During the inference phase, the testing dataset differs in both area and viewpoints from the training sets, and it comprises 1) real images from the UAVD4L region and 2) real images from the Swiss-EPFL region.
The input image size for the segmentation module remains \( (1024, 1024) \), and the $\delta_s$ is set to 0.5. 
In the pose selection stage, the sizes of the query silhouettes and rendered candidates are set to \( (301, 224) \) to capture approximate location information. The sampling interval is set to 10 meters and $\delta_I$ is defined as 127. In the fine pose estimation stage, the mask size is doubled to \( (602, 448) \).
We set the iteration count to \( N = 40 \), using 2 beams and an initial angular perturbation of 2 degrees. The translation perturbation follows a Gaussian distribution \( X \sim \mathcal{N}(0, \sigma^2) \) with a mean of 0 and \( \sigma = 1.5 \). An attenuation coefficient \( \gamma = 0.3 \), and \( n = 52 \) candidates are generated in each iteration.

Benefiting from the OSG~\cite{OSG} real-time rendering technique, the rendering speed can reach $630\mu s$ at a resolution of 602$\times$448. The entire training and testing process was conducted on 2 NVIDIA RTX 4090 GPUs. To validate the effectiveness of our module, we experimented with three model variants: (1) -no refine, which excludes the fine pose estimation stage; (2) -no select, which skips the coarse pose selection stage; and (3) Full model represents our proposed LoD-Loc v2 model.


\subsection{Evaluation Results}
\label{exp:UAV1_3}

\input{table/table3.tex}

\input{table/table2.tex}
\noindent \textbf{Evaluation over UAVD4L-LoDv2 dataset.} 
As described in \cref{Tab:table_1}, our approach performs excellently in both \textit{in-Traj.} and \textit{out-of-Traj.} queries. While UAVD4L demonstrates impressive performance, it still encounters challenges in retrieval that result in inaccuracies in the estimated poses~\cite{zhu2024lod}. We further compare with CAD-Loc, MC-Loc and LoD-Loc, which leverage the same 3D reference as ours. However, CAD-Loc, and MC-Loc struggle with localization due to the large modality gap between textured and non-textured images.  
LoD-Loc achieves better results by leveraging line features, but the limited line information in LoD1 models directly impacts localization accuracy. 
\cref{fig:exp_seg} shows the segmentation results and visualizations of each iteration. 


\noindent \textbf{Evaluation over Swiss-EPFLv2 dataset.} 
\cref{Tab:table_swiss} presents the inference results on the Swiss-EPFLv2 dataset. Our method is competitive for Swiss-EPFL scenes, despite being trained on UAVD4L region only. Apart from the metrics of \textit{out-of-Place} sequence, which is lower than those in the UAVD4L~\cite{wu2024uavd4l}, all other metrics surpass SOTA baselines by a large margin. Note that the comparison with texture model-based baselines is quite unfair, as these baselines use high-precision texture models enriched with detailed texture and geometry as references. 

\noindent \textbf{Evaluation over high-LoD dataset.} 
We conducted experiments on datasets with different levels of detail, including LoD3 (UAVD4L-LoD) and LoD2 (Swiss-EPFL), to validate the generalizability of our algorithm. \cref{Tab:table_2} demonstrate that LoD-Loc v2 achieves robust and accurate localization across datasets with varying LoD levels.

\input{table/table4.tex}

\subsection{Ablation Study}
\label{exp:ab_study}
We conduct ablation studies on the UAVD4L-LoDv2, focusing on convergence basin and segmentation modules. More ablation studies are provided in the Appendix.

\begin{figure}[t]
	\centering
	\begin{subfigure}{0.48\linewidth}
		\centering
		\includegraphics[width=1\linewidth]{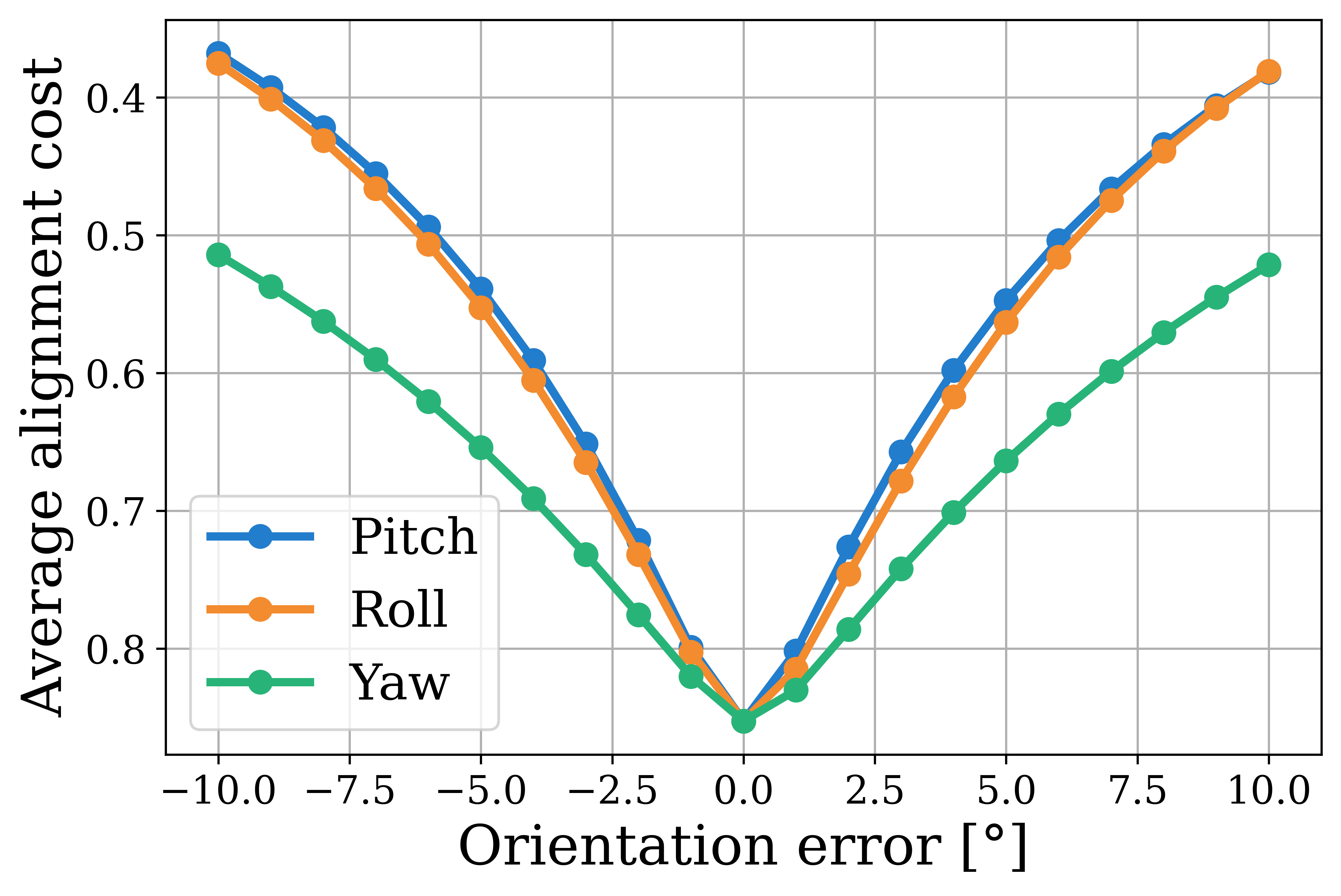}
	\end{subfigure}
	\centering
	\begin{subfigure}{0.466\linewidth}
		\centering
		\includegraphics[width=1\linewidth]{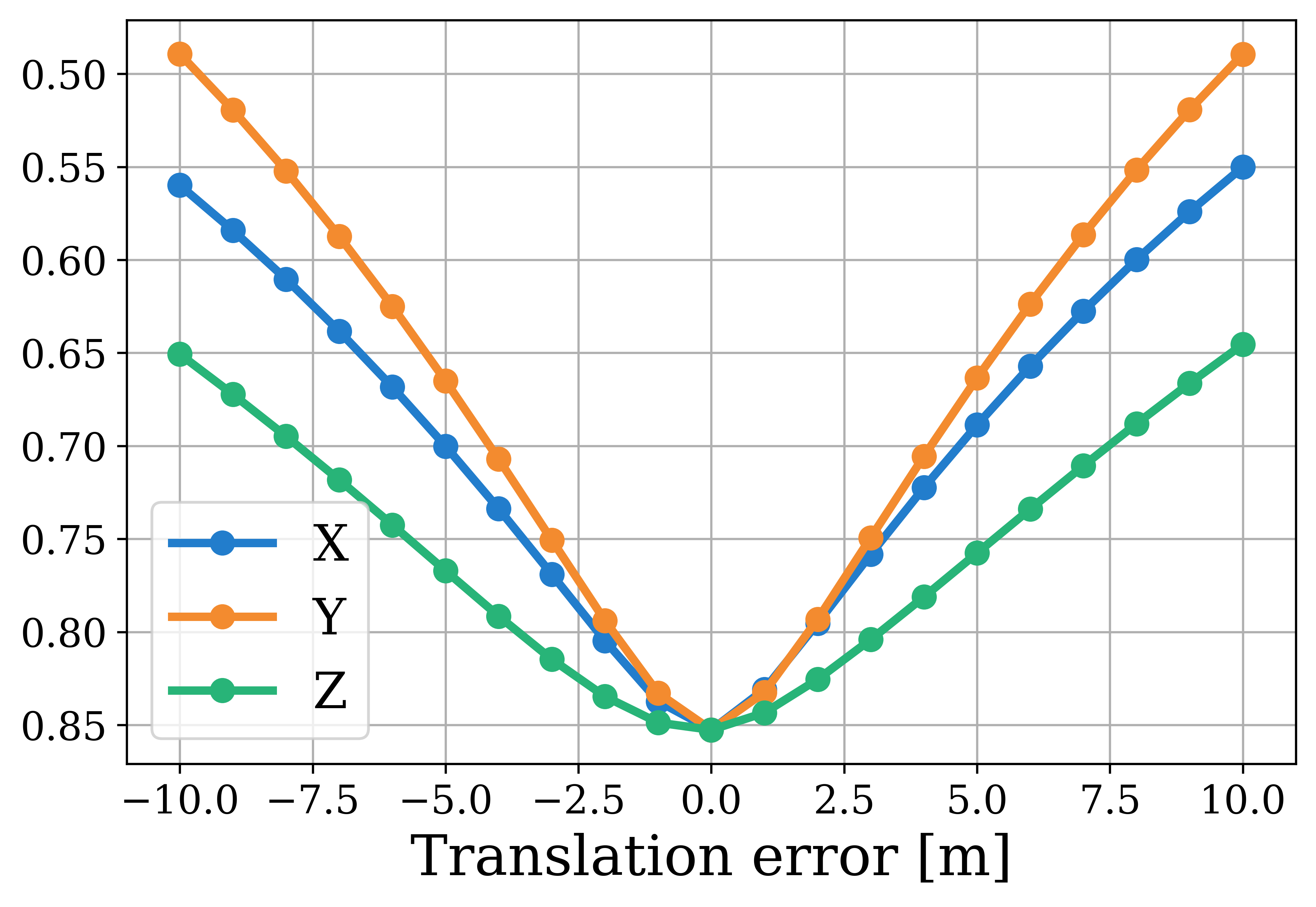}
	\end{subfigure}
 \caption{\textbf{Convergence basin concerning translation and orientation.} The average alignment cost is computed on the UAVD4L-LoDv2 dataset using \cref{eq:cost}.}
 \label{Fig:converge_basin}
\end{figure}

\noindent \textbf{Convergence basin.} 
We initialize different prior error ranges on the UAVD4L-LoDv2 dataset to explore the convergence basin of the proposed method. The X-Y-Z axis error ranges are set to \(\Delta = \pm [30, 50, 100, 200]\) meters. As shown in \cref{Tab:table_3}, although the accuracy of LoD-Loc v2 decreases slightly with increasing prior error, it still outperforms LoD-Loc by a large margin.
These results demonstrate that LoD-Loc v2 expands the convergence basin to accommodate larger prior errors, indicating its potential ability to solve the UAV localization problem even under GPS-constrained conditions. \cref{Fig:converge_basin} visualizes the convergence basin with respect to translation and orientation.

\noindent \textbf{Segmentation modules.} 
As shown in \cref{Tab:table_4}, we used two additional segmentation models, FastSAM~\cite{zhao2023fast} and SAM1-Adapter~\cite{chen2023sam}, as backbones for our ablation studies. The results demonstrate that the U-Net segmentation module with the SAM2 encoder achieves superior performance compared to other models.

%% file: table/table_swiss.tex
\begin{table*}[!t]
\centering
\setlength{\tabcolsep}{8.5pt} 
\begin{tabular}{lccccccccc} 
\toprule
\multicolumn{2}{c}{\multirow{2}{*}{Method}}                 & \multicolumn{4}{c} {\textit{in-Place.}}                    & \multicolumn{4}{c}{\textit{out-of-Place.}}                 \\ \cmidrule(r){3-6} \cmidrule(r){7-10} 
\multicolumn{2}{c}{}                                        & 2m-2°           & 3m-3°           & 5m-5°     & T.e./R.e.      & 2m-2°           & 3m-3°           & 5m-5°     & T.e./R.e.      \\ \cmidrule(r){1-10}
\multicolumn{2}{c}{Prior}                            & 0              & 0              &  0.56            & 17.6/3.87              & 0      &0
 &1.06   &17.9/3.94                    \\ \hline
\rowcolor[rgb]{1,0.992,0.980} {\cellcolor[rgb]{1,0.992,0.980}}                          & SIFT+NN        & 15.17 & 23.74 & 35.11  &2.57/1.54     & 32.98 & 54.35 &71.50   &  2.76/1.59                                      \\
\rowcolor[rgb]{1,0.992,0.980} {\cellcolor[rgb]{1,0.992,0.980}}                          & SPP+SPG                     & 33.85 & 56.32 & 72.75 & 2.57/1.54      & 77.04 & 89.71 & 92.35 &1.12/1.17                                        \\
\rowcolor[rgb]{1,0.992,0.980} {\cellcolor[rgb]{1,0.992,0.980}}                          & LoFTR                     & 26.40 & 46.21 & 62.22   &3.06/1.86    & 68.87 & 81.00 & 84.96  &1.12/1.08                                        \\
\rowcolor[rgb]{1,0.992,0.980} {\cellcolor[rgb]{1,0.992,0.980}}                          & e-LoFTR                    & 37.64 & 60.96 & 76.40  & 2.38/1.45     & 81.53 & 91.03 & 93.93 &0.91/1.08                                        \\
\rowcolor[rgb]{1,0.992,0.980} \multirow{-5}{*}{{\cellcolor[rgb]{1,0.992,0.980}}\begin{tabular}[c]{@{}>{\cellcolor[rgb]{1,0.992,0.980}}c@{}}UAVD4L~\cite{wu2024uavd4l}\\ \textit{Texture model}\end{tabular}} & RoMA                      & 45.95 & 66.77 & 80.73   &2.08/1.29    & \textbf{89.18} & \textbf{89.68} & \textbf{98.84}  & \textbf{0.77}/1.04                                        \\  \hline

\rowcolor[rgb]{0.941,0.941,0.941} \begin{tabular}[c]{@{}>{\cellcolor[rgb]{0.941,0.941,0.941}}c@{}}CAD-Loc~\cite{panek2023visual}\\\textit{LoD model}\end{tabular} & \textit{same*}                                                           & $\boldsymbol{0}$                                & $\boldsymbol{0}$                                & $\boldsymbol{0}$                &-                 & $\boldsymbol{0}$                                 & $\boldsymbol{0}$       &$\boldsymbol{0}$                                & -                    \\ 
\hline
\rowcolor[rgb]{0.902,0.902,0.902} {\cellcolor[rgb]{0.902,0.902,0.902}}                                                                                                                    & DINOv2                      & 0.90                                         & 4.40                                        & 17.50           &8.18/2.54                               & 2.90                                         & 9.00   & 30.20   &6.23/1.97                                                                                                                    \\
\rowcolor[rgb]{0.902,0.902,0.902} \multirow{-2}{*}{{\cellcolor[rgb]{0.902,0.902,0.902}}\begin{tabular}[c]{@{}>{\cellcolor[rgb]{0.902,0.902,0.902}}c@{}}MC-Loc~\cite{trivigno2024unreasonable}\\\textit{LoD model}\end{tabular}} & RoMa                     & 0.20                                         & 1.20                                         & 4.80          &9.80/2.65                                & 0.70                                         & 2.10   & 11.5                    &10.3/3.97                                                      \\  \hline
\rowcolor[rgb]{0.851,0.851,0.851} \begin{tabular}[c]{@{}>{\cellcolor[rgb]{0.851,0.851,0.851}}c@{}}LoD-Loc~\cite{zhu2024lod}\\\textit{LoD model}\end{tabular} & -                                                           & 36.79                                 & 50.56                                 & 69.77        &2.87/1.78                         & 14.24                                 & 31.39   &       59.89      & 8.73/2.78                                                  \\ 
\hline
\rowcolor[rgb]{0.898,0.898,1} {\cellcolor[rgb]{0.898,0.898,1}}                                                                                                                          & no refine              & 0.56           & 3.73           & 20.79   &7.37/3.76       & 0.53          & 2.11     &11.35   &8.92/3.90           \\
\rowcolor[rgb]{0.898,0.898,1} {\cellcolor[rgb]{0.898,0.898,1}}                                                                                                                          & no select              & 52.10          & 72.10          & 88.30   &1.90/0.89       & 31.10         & 55.90  &    81.30    &2.73/0.73             \\
\rowcolor[rgb]{0.898,0.898,1} \multirow{-3}{*}{{\cellcolor[rgb]{0.898,0.898,1}}\begin{tabular}[c]{@{}>{\cellcolor[rgb]{0.898,0.898,1}}c@{}}\textbf{LoD-Loc v2}\\\textit{LoD model}\end{tabular}} & \textbf{Full}          & \textbf{54.20} & \textbf{74.60} & \textbf{92.00} &\textbf{1.83/0.85} & 31.40 & 58.53   & 86.30 &2.64/\textbf{0.73} \\
\hline
\end{tabular}
\caption{\textbf{Quantitative comparison results of different methods over Swiss-EPFLv2 dataset.} The \textit{same*} indicates that the variants are identical to those in \cref{Tab:table_1}. T.e. and R.e. have the same meanings as those in \cref{Tab:table_1}.}
\label{Tab:table_swiss}
\end{table*}

%% file: table/table3.tex
\begin{table}
\centering
\begin{tabular}{l@{\hskip 0.2cm}c|c@{\hskip 0.1cm}c} 
\toprule
\multirow{2}{*}{Model}      & \multirow{2}{*}{\textbf{$\Delta$}} & \textit{in-Traj.}          & \textit{out-of-Traj.}                        \\  
\cline{3-4}
                        &                            & \multicolumn{2}{c}{( 2m, 2$^\circ$ ) / ( 3m, 3$^\circ$ ) / ( 5m, 5$^\circ$ )}  \\ 
\midrule
\multirow{4}{*}{LoD-Loc}    & 30                         & 6.86 / 12.9 / 20.9~ & 8.12 / 15.4 / 25.6          \\
                            & 50                         & 1.75 / 4.24 / 7.04~ & 2.87 / 4.88 / 8.07             \\
                            & 100                        & 0.31 / 1.00 / 1.81~ & 1.09 / 1.78 / 2.74                     \\ 
                             & 200                        & 0.12 / 0.19 / 0.31~ & 0.32 / 0.41 / 0.55 \\
\hline
\multirow{4}{*}{\makecell[c]{\textbf{LoD-Loc}\\ \textbf{v2}}} & 30                         & 92.6 / 97.1 / 98.5~ & 96.6 / 99.0 / 99.1          \\
                            & 50                         & 91.3 / 95.1 / 96.7~ & 95.9 / 98.3 / 98.5          \\
                            & 100                        & 91.3 / 95.1 / 96.2~ & 96.8 / 98.1 / 98.1          \\
                            & 200                        & 90.3 / 94.1 / 95.2~ & 94.7 / 97.2 / 97.4  \\
\bottomrule
\end{tabular}
\caption{\textbf{Performance under larger prior errors.} We present results for different prior poses with added translation errors, while an analysis of orientation errors is provided in the Appendix.}
\label{Tab:table_3}
\end{table}

%% file: table/table2.tex
\begin{table*}[!t]
\centering
\setlength{\tabcolsep}{13.8pt} 
\begin{tabular}{cccccccc} 
\toprule
\multirow{2}{*}{Dataset}                                                         & \multirow{2}{*}{Method} & \multicolumn{3}{c}{\textit{in-Place.} (\textit{in-Traj.})}                     & \multicolumn{3}{c}{\textit{out-of-Place.} (\textit{out-of-Traj.})}                     \\ 
\cmidrule(r){3-5} \cmidrule(r){6-8} 
             &                        & 2m-2$^\circ$           & 3m-3°           & 5m-5°            & 2m-2°           & 3m-3°           & 5m-5°             \\ 
\hline                                                                                                                   
LoD2         & LoD-Loc             & 48.60          & 65.31          & 79.78          & 37.73          & 57.26          & 77.57           \\
(Swiss-EPFL) & \textbf{LoD-Loc v2} & \textbf{57.90} & \textbf{76.00} & \textbf{90.60} & \textbf{40.40} & \textbf{63.90} & \textbf{85.80}                                         \\
\hline
LoD3 & LoD-Loc                      & 84.41                                         & 91.77                                         & 96.95                                         & 95.94                                         & 99.00                                       & 99.36                                           \\
(UAVD4L-LoD) & \textbf{LoD-Loc v2}                     & \textbf{98.50}                                         & \textbf{99.60}                                         & \textbf{99.80}                                         & \textbf{99.40}                                         & \textbf{99.80}                                        & \textbf{100.00}                                        \\
\hline
\end{tabular}
\caption{\textbf{Quantitative comparison results of different methods over LoD3 and LoD2 dataset.} Compared to the LoD-Loc, our approach significantly improves visual localization performance on high-LoD models.}
\label{Tab:table_2}
\end{table*}

%% file: table/table4.tex
\begin{table*}[t]
\setlength{\tabcolsep}{13.8pt} 
\centering
\begin{tabular}{lcccccccc} 
\toprule
\multirow{2}{*}{Method} & \multicolumn{4}{c}{\textit{in-Traj.}} & \multicolumn{4}{c}{\textit{out-of-Traj.}}  \\ 
\cmidrule(r){2-5} \cmidrule(r){6-9}
                           & 2m-2$^\circ$ & 3m-3$^\circ$ & 5m-5$^\circ$ & IoU & 2m-2$^\circ$ & 3m-3$^\circ$ & 5m-5$^\circ$ & IoU  \\ 
\hline
FastSAM                    & 45.3   & 69.8   & 90.1 &0.39   & 45.6  & 67.2   & 87.4 & 0.31         \\
SAM1-Ada.               & 86.4   & 94.7   & 98.6 & 0.54  & 96.4   & 97.8   & 98.3  &0.43        \\ \hline
\textbf{Ours}                  & \textbf{93.7}   & \textbf{98.4}   & \textbf{99.5} &\textbf{0.88}  & \textbf{97.9}   & \textbf{99.8}  & \textbf{100.0}   &\textbf{0.80}      \\
\bottomrule
\end{tabular}
\caption{\textbf{Ablation study on different segmentation modules.} Note that each variant is fine-tuned separately, using consistent training data and input image sizes, while we keep other parameters at default settings.}
\label{Tab:table_4}
\end{table*}

%% file: sec/5_Conclusion.tex
\section{Conclusion}
\label{sec: Conclusion}

This paper introduces LoD-Loc v2, a novel method for aerial visual localization using low-LoD city models. Given a build-in prior pose data, LoD-Loc v2 employs a coarse-to-fine pipeline to recover the camera pose, consisting of building silhouette segmentation, coarse pose selection, and fine pose estimation. Furthermore, we contribute two novel datasets with model level of LoD1, along with real query images and ground-truth pose annotations. 
LoD-Loc v2 achieves excellent performance both on high-LoD and low-LoD datasets, even surpassing recent SOTA texture model-based localization techniques. Moreover, our proposed method extends the convergence basin to accommodate larger prior errors, making it a potential solution for GPS-constrained scenarios.
We believe that LoD-Loc v2 opens new possibilities for global aerial visual localization by utilizing widely available low-LoD city models.


\noindent \textbf{Limitation.} Our method encounters limitations when building silhouettes predominantly occupying the query image. However, this situation is relatively uncommon in the context of UAV localization, where the capturing distance to buildings is typically substantial. Further analysis and visualizations are provided in the Appendix.

